\documentclass[lettersize,journal]{IEEEtran}
\usepackage{amsmath,amsfonts}
\usepackage{algorithmic}
\usepackage{algorithm}
\usepackage{array}
\usepackage[caption=false,font=normalsize,labelfont=sf,textfont=sf]{subfig}
\usepackage{textcomp}
\usepackage{stfloats}
\usepackage{url}
\usepackage{verbatim}
\usepackage{graphicx}
\usepackage{cite}
\usepackage{graphicx}
\usepackage{amsmath}
\usepackage{algorithm}
\usepackage{algorithmic}
\usepackage{float}

\usepackage[utf8]{inputenc} 
\usepackage[T1]{fontenc}    
\usepackage{hyperref}       
\usepackage{url}            
\usepackage{booktabs}       
\usepackage{amsfonts}       
\usepackage{nicefrac}       
\usepackage{microtype}      
\usepackage{xcolor}         
\usepackage{dirtytalk}
\usepackage{rotating}

\usepackage{adjustbox} 
\usepackage{booktabs}
\usepackage{multirow}

\begin{document}

\title{Data Augmentation for Deep Learning Regression Tasks by Machine Learning Models}

\author{%
  Assaf Shmuel \\ 
  Department of Computer Science, Bar Ilan University\\
   Oren Glickman \\
  Department of Computer Science, Bar Ilan University\\
    Teddy Lazebnik \\
  Department of Cancer Biology, Cancer Institute, UCL\\
}

\markboth{}%
{}

\maketitle

\begin{abstract}
Deep learning (DL) models have gained prominence in domains such as computer vision and natural language processing but remain underutilized for regression tasks involving tabular data. In these cases, traditional machine learning (ML) models often outperform DL models. In this study, we propose and evaluate various data augmentation (DA) techniques to improve the performance of DL models for tabular data regression tasks. We compare the performance gain of Neural Networks by different DA strategies ranging from a naive method of duplicating existing observations and adding noise to a more sophisticated DA strategy that preserves the underlying statistical relationship in the data. Our analysis demonstrates that the advanced DA method significantly improves DL model performance across multiple datasets and regression tasks, resulting in an average performance increase of over 10\% compared to baseline models without augmentation. The efficacy of these DA strategies was rigorously validated across 30 distinct datasets, with multiple iterations and evaluations using three different automated deep learning (AutoDL) frameworks: AutoKeras, H2O, and AutoGluon. This study demonstrates that by leveraging advanced DA techniques, DL models can realize their full potential in regression tasks, thereby contributing to broader adoption and enhanced performance in practical applications.\end{abstract}

\begin{IEEEkeywords}
data augmentation, machine learning, deep learning.
\end{IEEEkeywords}

\section{Introduction}
\label{sec:introduction}
Deep learning (DL) has achieved remarkable success in various domains such as computer vision \cite{cv_example,cv_2,lastcite_2}, signal processing \cite{signal_example_1,signal_example_2}, and natural language processing \cite{nlp_example,nlp_example_2}. However, when applied to tabular data, which represents one of the most simplistic forms of data, DL models often underperform in comparison to classical machine learning (ML) models \cite{dl_table_shit_1,dl_table_shit_2}. This performance gap can be attributed, in part, to the inherent characteristics of tabular data, which typically consists of heterogeneous, low-dimensional features with varied scales and distributions. 

DL models excel in high-dimensional, structured data where they can leverage their deep architectures to learn hierarchical feature representations. However, when applied to tabular data, DL models struggle due to their tendency to overfit and their requirement for large amounts of data to achieve comparable performance \cite{tabular_issue}. Moreover, DL models often require extensive hyperparameter tuning and sophisticated architecture design, which adds to their complexity without necessarily translating into better performance for tabular datasets. 

One strategy to address these limitations is applying data augmentation (DA), a technique that artificially expands training datasets \cite{review_intro_1}. In the context of tabular data regression, DA aims to generate new synthetic data points that preserve the underlying statistical properties of the original data, thereby improving model generalizability and performance \cite{review_intro_2,review_intro_3}. Various DA techniques exist, each based on the assumption that minor changes, consistencies, or symmetries applied to the input will not alter the assigned class label \cite{intro_insp_1}. These methods introduce novel variations as modified versions of existing training examples, adjusting the input's distribution, and enhancing the algorithm's resilience to scenarios where the test set's distribution may vary from that of the training set \cite{intro_insp_2}.

Despite the potential of DA in the context of tabular data for DL models, this method is mostly unattended. Two prominent studies that have applied DA to tabular data regression are the Mixup augmentation \cite{mixup} and Anchor Data Augmentation \cite{base_paper}. Mixup augmentation presents a straightforward and universal method of data augmentation by employing convex combinations of samples and was demonstrated on tabular data classification tasks. A recent extension of the Mixup algorithm, C-Mixup, adjusts the sampling probability based on the similarity of the labels rather than selecting training examples for mixing with uniform probability as performed by Mixup \cite{c-mixup}. Anchor Data Augmentation \cite{base_paper} employs multiple duplicates of altered samples within anchor regression \cite{ancor_regression} to provide additional training instances, resulting in more resilient regression predictions. 

These works highlight the challenge of generating both new samples in the source space (i.e., \(X\)) and their labels (i.e., \(y\)) for regression tasks. This challenge is less prominent in classification tasks where more often than not, small changes in the source space often do not alter the label. Consequently, methods attempting to perform DA by jointly predicting both source and target usually generate results extremely close to the original data to avoid this issue \cite{issue_1,issue_2}. 

In this study, we investigate the effectiveness of various DA techniques for tabular regression tasks. Specifically, we explore a range of methods, from relatively simple techniques like adding noise to existing data points to more sophisticated ones that leverage automated machine learning frameworks to discover optimal augmentation pipelines. We evaluate the impact of these techniques on model performance across 30 well-known datasets. Our results show over 10\% improvement, on average, compared to performing DL on the tabular data without DA. We compare our method to existing state of the art augmentation methods and find it to have significantly higher performance.

The rest of the paper is organized as follows. Section \ref{sec:related_work} describes state-of-the-art DA methods and their adoption for tabular data as well as AutoDL methods. Section \ref{sec:methodology} formally introduces the proposed DA for DL using ML. Section \ref{sec:experiments} outlines the experimental framework we used to evaluate the proposed method. Section \ref{sec:results} shows the obtained results. Section \ref{sec:discussion} discusses the possible applications of our results with their limitations and suggests possible future work. 

\section{Related Work}
\label{sec:related_work}
In this section, we provide a quick overview of data augmentation methods as well as the current state-of-the-art in the field of AutoDL for regression tasks. 

\subsection{Data augmentation}
Most of the DA methods can be associated with image, textural, and time-series (signal) datasets which sometimes can be adapted to tabular data \cite{new_rw_3,new_rw_2,new_rw_1}.   
For example, In AutoAugment \cite{autoaugment} a neural network undergoes training via Reinforcement Learning to amalgamate a range of transformations with different intensities for application on samples from a specified dataset, aiming to enhance the model's accuracy. This method has been originally proposed and examined for image data rather than tabular data but can be easily adopted as vectorization of the image is used at the beginning of the process. Further extensions such as RandAugment \cite{new_rw_7}, Online Hyperparameter Learning \cite{new_rw_8}, and DivAug \cite{new_rw_9}; explore reducing the cost of the pertaining search phase with randomized transformations, using multiple data augmentation workers that are updated using evolutionary strategies and Reinforcement Learning, and finds the policy which maximizes the diversity of the augmented data, respectively, in order to further improve the DA for \say{down the stream} regression and classification performance.  

In addition, dedicated tabular DA methods have been introduced to a wide range of tasks \cite{new_rw_5,rw_start_1}. Some relatively simple methods involve perturbing existing data points by adding random noise \cite{da_noise}. However, these methods can introduce irrelevant information and potentially harm model performance \cite{da_noise_bad_idea}. More sophisticated methods leverage techniques like random swapping of categorical feature values \cite{da_smarter} or smoothing techniques to modify existing data points while maintaining the overall data distribution \cite{da_smarter_review}. These methods have shown some promise, but their effectiveness can be limited, particularly for complex datasets. 

To tackle this challenge, methods utilizing sophisticated data-driven methods are proposed for the DA task itself \cite{og1,og2, og3,og4}. For instance, \cite{new_rw_4} introduce a DA method specifically tailored for tabular data, which is applied to the semi-supervised learning algorithm. This method is based on a multi-task learning framework comprising two main components: the data augmentation process and the consistency training process. The data augmentation process involves perturbing the data in latent space using a variational auto-encoder (VAE) to generate augmented samples as reconstructed samples. \cite{new_rw_6} proposed the Histogram Augmentation Technique that generates samples that preserve the distribution of each feature in the dataset independently. The authors show this method has desired properties such as not affecting much the feature importance of ML classifiers while also improving on average the classifier's accuracy. 

\subsection{Automatic Deep Learning}
Data-driven models such as ML and DL applications generally involve several steps such as data preparation, feature engineering, algorithm selection, and hyperparameter tuning \cite{ml_breif_intro,pipeline_done_1,expriment_idea}. Many of these steps involve trial and error, particularly for those who are not ML/DL experts \cite{substract,yakovlev2020oracle,mu2021assassin}. Experienced practitioners often rely on heuristics to navigate the vast dimensional space of parameters \cite{teddy_2,fs_rw_1,fs_rw_2}. As the number of non-specialists working with ML has grown, there has been a recent push to automate various components of the ML workflow. This has led to the development of Automated Machine Learning (AutoML) and Automated Deep Learning (AutoDL) \cite{algo_choose_3,lastcite_1,algo_choose_1,auto_sklearn,pinto}. 

While multiple models are proposed for automatic DL \cite{hyper_find_1,he2021automl}, one can name three popular methods with high adoption over a wide range of applications -  AutoKeras \cite{autokeras}, H2O \cite{h2o}, and AutoGluon \cite{autogluon}. AutoKeras is an open-source AutoML library that is built on top of Keras and TensorFlow, and aims to make DL accessible to everyone. AutoKeras automates the model selection and hyperparameter tuning processes, enabling users to build high-performing neural networks with minimal effort. It provides easy-to-use interfaces for tasks such as image classification, text classification, and regression \cite{autokeras}. H2O is a comprehensive open-source ML platform that includes H2O AutoML, a tool for automating the training and tuning of ML models. H2O AutoML supports a wide range of algorithms, including DL, gradient boosting machines, and generalized linear models. H2O's DL capabilities are built on a multi-layer feedforward artificial neural network framework, which includes advanced features like adaptive learning rates and regularization techniques. AutoGluon is another powerful open-source AutoML library developed by AWS Labs. It aims to make ML more accessible by providing a simple interface for training and deploying high-quality models. AutoGluon supports a variety of ML tasks, including tabular data prediction, text and image classification, and object detection. AutoGluon automates model selection, hyperparameter optimization, and ensembling, making it an effective tool for both beginners and experts in ML \cite{autogluon}. 

\section{Data Augmentation for Deep Learning Regression Using Machine Learning}
\label{sec:methodology}
\begin{figure*}[!ht]
    \centering
\includegraphics[width=0.99\textwidth]{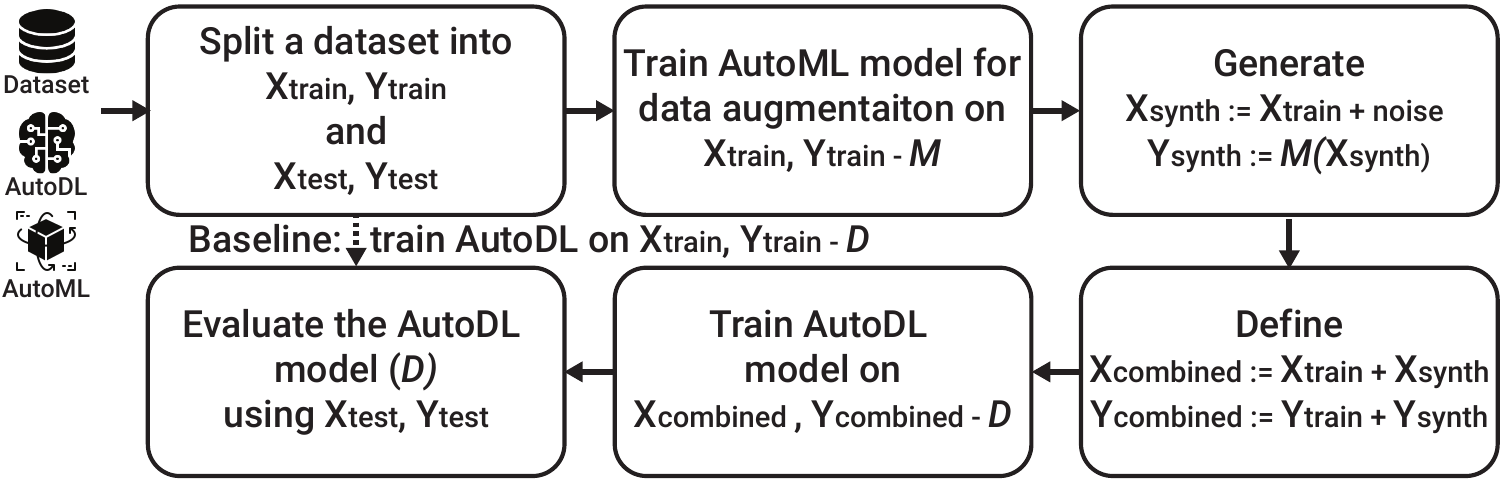}
    \caption{A schematic view of the method's flow. }
    \label{fig:scheme}
\end{figure*}

In this section, we outline the proposed DL regression model using ML models as DA. First, we formally introduce the proposed method. Next, we describe the evaluation strategy and the experimental setup used for this study. Finally, we provide a description of the real-world datasets used for the experiments. Figure \ref{fig:scheme} presents a schematic view of the study's flow. 

The core idea of the proposed method is to improve the performance of DL models by generating synthetic data and leveraging ML models to assign labels to this data. Unlike previous studies that primarily used clustering methods for label evaluation, we treat this as a classical ML task: given $X_{synth}$, we use ML models to evaluate $y_{synth}$ based on the patterns learned from $(X_{train},y_{train})$. The steps involved are as follows:

First, an AutoML is trained on the original training dataset (\(X_{train}\) and \(y_{train}\)) producing a model $M$. The model is not restricted to being a DL model; any ML model capable of generating additional data that reflects the original dataset's dynamics can be employed. By running AutoML we, hopefully, obtain the best ML model for the given dataset. 
Next, additional synthetic observations (\(X_{synth}\)) are generated. These synthetic features are derived from the existing training data \(X_{\text{train}}\) by sampling observations and incorporating noise. For each sampled observation, we add noise to all feature columns by computing the mean \(\mu_c\) and standard deviation \(\sigma_c\) for each column \(c\). The noise is generated from a normal distribution \(\mathcal{N}(\mu_c, \eta \cdot \sigma_c)\), where \(\eta\) is a scaling factor that determines the magnitude of the noise relative to the standard deviation.
Using the trained AutoML model $M$, we predict the corresponding synthetic responses (\(y_{synth}\)) for these newly generated features (\(X_{synth}\)). 
The original training data (\(X_{train}\) and \(y_{train}\)) is augmented with the synthetic data (\(X_{synth}\) and \(y_{synth}\)) to form combined datasets \(X_{combined}\) and \(y_{combined}\).
With the augmented training dataset (\(X_{combined}\) and \(y_{combined}\)), we train an AutoDL model $D$.

A pseudocode of this method is presented in Algorithm 1.

\begin{algorithm}[H]
\caption{Data Augmentation for DL Regression using an AutoDL Model}
\label{alg:da_dl}
\begin{algorithmic}[1]
    \REQUIRE Dataset $(X, y)$, AutoDL model $)\mathcal{L}($, AutoML model $)\mathcal{M})$, Augmentation volume $(V)$, Noise fraction $(\eta)$ \\
    \STATE \textbf{Output:} Performance metrics $(P_{\text{baseline}})$ and $(P_{\text{aug}})$
    
    \STATE Split $(X, y)$ into $(X_{\text{train}}, y_{\text{train}})$ and $(X_{\text{test}}, y_{\text{test}})$
    \STATE $P_{\text{baseline}} \gets$ Call \textbf{BaselinePerformance} with $(X_{\text{train}}, y_{\text{train}}, X_{\text{test}}, y_{\text{test}}, \mathcal{L})$
    \STATE $(X_{\text{synth}}, y_{\text{synth}}) \gets$ Call \textbf{GenerateSyntheticData} with $(X_{\text{train}}, y_{\text{train}}, V, \eta, \mathcal{M})$
    \STATE $X_{\text{combined}} \gets X_{\text{train}} \cup X_{\text{synth}}$
    \STATE $y_{\text{combined}} \gets y_{\text{train}} \cup y_{\text{synth}}$
    \STATE $P_{\text{aug}} \gets$ Call \textbf{TrainAugmentedDLModel} with $(X_{\text{combined}}, y_{\text{combined}}, X_{\text{test}}, y_{\text{test}}, \mathcal{L})$
    \STATE \textbf{return} Performance metrics $P_{\text{baseline}}$ and $P_{\text{aug}}$ 
    
    \STATE \textbf{BaselinePerformance}$(X_{\text{train}}, y_{\text{train}}, X_{\text{test}}, y_{\text{test}}, \mathcal{L})$:
        \STATE Train $\mathcal{L}$ on $(X_{\text{train}}, y_{\text{train}})$
        \STATE Evaluate $\mathcal{L}$ on $(X_{\text{test}}, y_{\text{test}})$
        \STATE Record baseline performance $P_{\text{baseline}}$
        \STATE \textbf{return} $P_{\text{baseline}}$

    \STATE \textbf{GenerateSyntheticData}$(X_{\text{train}}, y_{\text{train}}, V, \eta, \mathcal{M})$:
        \STATE Train AutoML model $\mathcal{M}$ on $(X_{\text{train}}, y_{\text{train}})$
        \STATE Randomly select $V$ samples from $X_{\text{train}}$ to form $X_{\text{synth}}$
        \FOR{each column $c$ in $X_{\text{train}}$}
            \STATE Calculate mean $\mu_c$ and standard deviation $\sigma_c$ of column $c$
            \FOR{each sample $s$ in $X_{\text{synth}}$}
                \STATE Add Gaussian noise to $s[c]$ with $\mathcal{N}(\mu_c, \eta \cdot \sigma_c)$
            \ENDFOR
        \ENDFOR
        \STATE Predict $y_{\text{synth}}$ using $\mathcal{M}$ on $X_{\text{synth}}$
        \STATE \textbf{return} $(X_{\text{synth}}, y_{\text{synth}})$

    \STATE \textbf{TrainAugmentedDLModel}$(X_{\text{combined}}, y_{\text{combined}}, X_{\text{test}}, y_{\text{test}}, \mathcal{L})$:
        \STATE Train $\mathcal{L}$ on $(X_{\text{combined}}, y_{\text{combined}})$
        \STATE Evaluate $\mathcal{L}$ on $(X_{\text{test}}, y_{\text{test}})$
        \STATE Record augmented performance $P_{\text{aug}}$
        \STATE \textbf{return} $P_{\text{aug}}$
    
\end{algorithmic}
\end{algorithm}
\section{Experiments}
\label{sec:experiments}
In this section, we outline the experimental setup used to explore the proposed method, including the models and datasets used and the valuation strategy. 

\subsection{Models and data}
\label{subsec:models}
We use the popular Tree-based Pipeline Optimization Tool (TPOT) library as an AutoML tool for data augmentation \cite{tpot}. TPOT is an open-source library that automates the process of designing and optimizing ML pipelines. It uses genetic programming to explore a wide range of models and preprocessing steps, aiming to find the best pipeline for a given dataset.

In each run, we train the TPOT model for ten minutes using the library's default parameters. This approach allows TPOT to explore various combinations of ML algorithms and preprocessing techniques within the given time frame. By leveraging TPOT's automated feature engineering and model selection capabilities, we can generate robust synthetic data that resembles the original dataset, thereby enhancing the performance of our DL models.

We use three different AutoDL libraries to obtain baseline DL performance, and also to examine the DL performance after augmentation. We use the AutoKeras \cite{autokeras}, H2O \cite{h2o}, and AutoGluon \cite{autogluon} libraries. Although H2O runs additional ML models, we limit its configuration to only run DL models for the purpose of this study. We run Autogluon with a maximum run time of ten minutes. We run H2O and AutoKeras with a maximum of 200 epochs. All three models were run with the same configuration for both the original training data and over the augmented data. By utilizing these three AutoDL libraries, we can establish a comprehensive baseline for DL performance. Additionally, we can evaluate the impact of data augmentation on DL performance across different AutoDL frameworks. 

We use 20 datasets of various sizes to compare our method with additional augmentation methods. We later use ten additional large datasets to explore the learning curve of the DL models and the contribution of our method in each training size. The datasets are provided in the supplementary materials (Table \ref{tab:datasets}).

\subsection{Evaluation}
\label{subsec:evaluation}
We begin by comparing the efficacy of our method with 20 additional real-world datasets. These datasets are mostly small and include a randomized train-test split (80-20\%). We compare the performance of the three AutoDL models either with or without augmentation; we repeat this process with our method, C-mixup \cite{c-mixup}, ADA \cite{base_paper}, a naive augmentation by adding noise to existing X rows without changing their labels, and a baseline without DA. We use 5\% noise, and perform a robustness test for 1\% and 10\%. For all augmentation methods, we augment 10000 rows, while later examining alternative values in robustness tests. We then seek to gain a deeper understanding of the effectiveness of our DA method, and to quantify how it is affected by the original training size and the number of augmented rows. For this purpose, we use ten large-scale datasets ($>10000$ observations, most $>50000$ observations), each time randomly sampling a different number of training rows (500, 1000, 5000, 10000, 50000). Each run is repeated ten times to ensure robustness. We evaluate our augmentation method on each of the ten datasets, for each of the five train sizes, with ten repetitions. To explore the effect of different augmentation volumes, we repeat this process five times with varying numbers of augmented rows (500, 1000, 5000, 10000, 50000). We summarize our findings by presenting the mean improvement compared to the baseline as a function of the original training size and of the augmented rows.

\section{Results}
\label{sec:results}

We begin by presenting the results of our method applied to 20 different datasets of varying sizes. Next, we analyze the improvement rate as a function of training size by examining 10 additional large datasets and sampling various training set sizes. Finally, we investigate whether the improvement in our method is purely due to knowledge distillation from the TPOT model, purely from data augmentation, or a combination of both.

\subsection{Comparison of various augmentation methods}
\label{subsec:twenty_datasets}

We examine the improvement by augmentation on 20 datasets of various sizes. Table \ref{tab:benchmark} presents a comparison between two state of the art augmentation methods (ADA \cite{base_paper} and C-mixup \cite{c-mixup}), a naive approach of augmentation by adding noise, a baseline without augmentation, and our method. We find a mean improvement of 12.07\% across all datasets compared to the baseline without DA (6.36\% for H2O, 18.35\% for AutoKeras, and 11.52\% for AutoGluon). While both ADA and C-mixup have a positive performance and outperform the baseline without augmentation, our method provides significantly higher performance. We obtained the best performance in 10 out of 20 datasets for H2O (second best are C-mixup and noise with 3), 9 out of 20 datasets for AK (second best is ADA with 4), and 7 out of 20 for AutoGluon (second best are C-mixup and noise with 5). The error bars of the table are presented in the Appendix, in Tables \ref{tab:h2o_errors}, \ref{tab:AK_errors}, and \ref{tab:gluon_errors}, and its statistical significance is analyzed in Table \ref{table:first_experiment_p_values}. We performed a robustness test in which we replaced the 5\% noise to generate X in our method with 1\% and 10\%, and found no significant differences.

\begin{table*}[!ht]
\caption{A comparison between our proposed DA model, ADA, C-mixup, naive noise, and baseline model with no DA. RMSE results reported over 20 datasets and 3 AutoDL models (H2O, AutoKeras, AutoGluo). The best augmentation method in each dataset (row) and with each AutoDL model (column) is in bold.}
\label{tab:benchmark}
\small
\centering
\hspace*{-0.08\textwidth}
\setlength{\tabcolsep}{1pt}
\begin{tabular}{@{}lccccc@{\hskip 8pt}ccccc@{\hskip 8pt}ccccc@{}}
\multirow{2}{*}{\centering DS} & \multicolumn{5}{c}{H2O} & \multicolumn{5}{c}{AutoKeras} & \multicolumn{5}{c}{AutoGluon} \\
\cmidrule(lr){2-6} \cmidrule(lr){7-11} \cmidrule(lr){12-16}
 & Ours & ADA & C-mixup & Noise & Baseline & Ours & ADA & C-mixup & Noise & Baseline & Ours & ADA & C-mixup & Noise & Baseline \\
\cmidrule(lr){2-6} \cmidrule(lr){7-11} \cmidrule(lr){12-16}
A & \textbf{4495} & 6055 & 5824 & 5963 & 5545 & \textbf{4444} & 5686 & 9513 & 6044 & 6517 & 4547 & 4659 & 4654 & \textbf{4326} & 4640 \\
B & \textbf{0.61} & 0.69 & 0.70 & 0.67 & 0.74 & \textbf{0.64} & 0.84 & 0.73 & 0.84 & 0.73 & \textbf{0.61} & 0.65 & 0.67 & 0.69 & 0.64 \\
C & \textbf{1.1} & 1.6 & 1.9 & 1.9 & 1.5 & \textbf{1.2} & 1.8 & 2.0 & 1.8 & 1.6 & \textbf{1.2} & 1.5 & 1.9 & 1.8 & 1.6 \\
D & \textbf{6.3} & 9.6 & 11.5 & 9.0 & 8.6 & \textbf{6.5} & 10.9 & 12.4 & 9.8 & 8.2 & \textbf{6.3} & 8.3 & 10.8 & 8.8 & 7.5 \\
E & \textbf{4.70} & 5.20 & 5.09 & 5.43 & 5.12 & \textbf{4.96} & 5.34 & 5.01 & 5.82 & 6.85 & \textbf{4.81} & 5.32 & 4.94 & 5.43 & 5.25 \\
F & 157 & 133 & 132 & \textbf{131} & 150 & 157 & 134 & 150 & \textbf{131} & 139 & 159 & 142 & \textbf{130} & 143 & 178 \\
G & 1.68 & \textbf{1.53} & 1.73 & 1.56 & 1.69 & 2.05 & 2.02 & 2.22 & \textbf{1.78} & 5.44 & 2.00 & 2.01 & 2.09 & \textbf{1.67} & 2.00 \\
H & \textbf{636} & 754 & 818 & 712 & 721 & \textbf{690} & 904 & 967 & 829 & 869 & \textbf{661} & 790 & 775 & 686 & 665 \\
I & \textbf{0.06} & 0.08 & 0.08 & 0.07 & 0.08 & \textbf{0.06} & 0.10 & 0.08 & 0.08 & 0.08 & \textbf{0.06} & 0.07 & 0.07 & 0.08 & 0.06 \\
J & 1.45 & 1.48 & 1.66 & \textbf{1.29} & 1.66 & 1.48 & 1.56 & 1.77 & 1.52 & \textbf{1.37} & 1.46 & 1.46 & 1.53 & \textbf{1.38} & 1.61 \\
K & 10.78 & 10.77 & 12.01 & \textbf{9.57} & 10.66 & 11.30 & 11.15 & 11.77 & \textbf{9.97} & 10.27 & 10.49 & 10.98 & 12.36 & \textbf{9.30} & 11.70 \\
L & 1.72 & \textbf{1.30} & 1.91 & 1.48 & 1.69 & 1.69 & 1.49 & 2.30 & 1.71 & \textbf{1.37} & 1.69 & 1.62 & \textbf{1.53} & 2.15 & 1.75 \\
M & \textbf{1.54} & 1.55 & 1.59 & 1.80 & 1.62 & 1.57 & \textbf{1.54} & 1.58 & 1.74 & 2.20 & 1.54 & 1.68 & 1.53 & 1.71 & \textbf{1.51} \\
N & 4.61 & 2.97 & \textbf{2.57} & 4.89 & 4.37 & 4.56 & 3.07 & \textbf{2.95} & 4.98 & 3.83 & 4.67 & 3.47 & \textbf{2.98} & 5.74 & 5.23 \\
O & 0.04 & 0.02 & \textbf{0.02} & 0.02 & 0.02 & 0.04 & 0.02 & \textbf{0.02} & 0.03 & 0.04 & 0.03 & 0.03 & 0.04 & \textbf{0.02} & 0.02 \\
P & 6.73 & 4.38 & \textbf{4.09} & 5.30 & 5.96 & 8.46 & \textbf{5.25} & 5.43 & 5.26 & 6.85 & 4.20 & \textbf{4.10} & 4.55 & 4.78 & 4.14 \\
Q & \textbf{139.93} & 186.06 & 152.95 & 164.97 & 159.00 & \textbf{152.79} & 214.75 & 189.53 & 180.89 & 167.69 & 128.03 & 130.53 & \textbf{125.50} & 158.32 & 165.29 \\
R & \textbf{2.24} & 3.08 & 3.45 & 2.96 & 3.17 & \textbf{2.60} & 3.68 & 3.82 & 3.31 & 3.47 & \textbf{2.46} & 3.18 & 3.55 & 3.15 & 3.35 \\
S & 0.13 & 0.09 & 0.16 & 0.18 & \textbf{0.09} & 0.13 & \textbf{0.08} & 0.14 & 0.21 & 0.10 & 0.13 & \textbf{0.10} & 0.14 & 0.13 & 0.21 \\
T & 0.30 & 0.32 & 0.51 & 0.48 & \textbf{0.26} & 0.31 & \textbf{0.30} & 0.40 & 0.54 & 0.33 & 0.37 & 0.37 & \textbf{0.34} & 0.48 & 0.43 \\
\cmidrule(lr){2-6} \cmidrule(lr){7-11} \cmidrule(lr){12-16}
\textbf{\# best} & \textbf{10} & \textbf{2} & \textbf{3} & \textbf{3} & \textbf{2} & \textbf{9} & \textbf{4} & \textbf{2} & \textbf{3} & \textbf{2} & \textbf{7} & \textbf{2} & \textbf{5} & \textbf{5} & \textbf{1} \\
\end{tabular}
\end{table*}

\subsection{Demonstration of the learning curve in large datasets}
\label{subsec:large_datasets}

\begin{figure*}[htp]
    \centering
    \hspace*{-0.15\textwidth}
    \includegraphics[width=18cm]{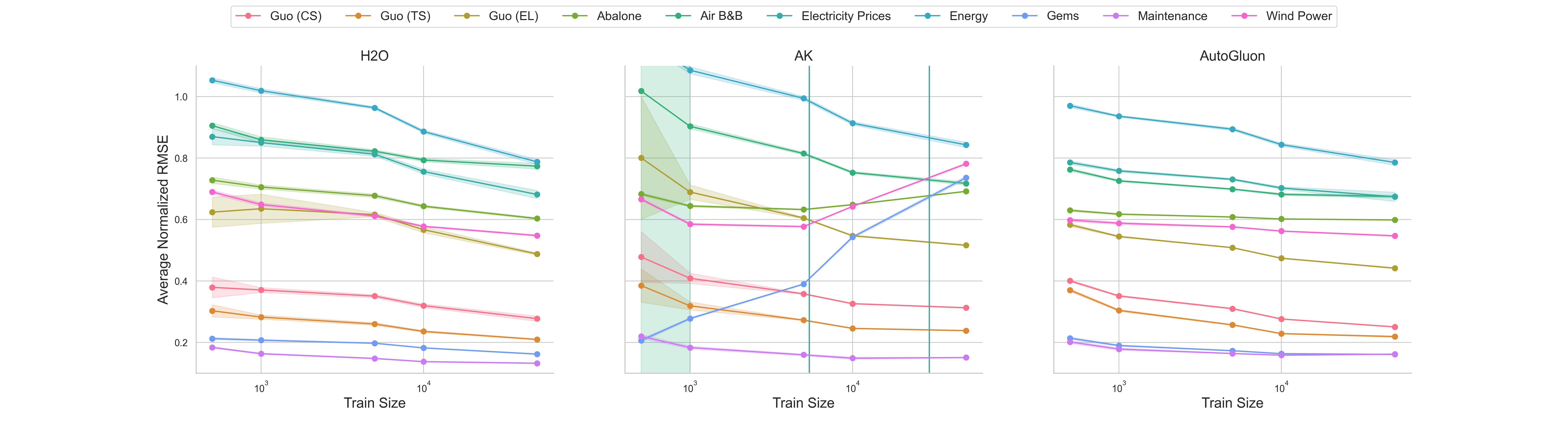}
    \caption{Normalized RMSE values as a function of training size. The figure presents a summary of 50 iterations for ten datasets without augmentation.}
    \label{fig:DL_results}
\end{figure*}

Figure \ref{fig:DL_results} presents the mean normalized RMSE for 50 iterations of each of the ten datasets. We demonstrate the improvement in DL performance across all three AutoDL libraries as the training size increases. The training size decreases performance in only one of 30 configurations (Gems with AK). While this result is to be expected, to the best of our knowledge it has not yet been shown in such extensive experiments. Specifically, by using the same datasets with varying training set sizes we eliminate differences that originate from comparisons of different datasets, and explicitly demonstrate the influence of size.

Table \ref{tab:large_datasets} presents the mean performance improvement obtained by augmentation across ten iterations in each of the ten datasets. The axes describe the baseline's training size (horizontal axis) and the number of augmented rows (vertical axis). As expected, the improvement obtained by augmentation is most crucial in small datasets, where the original data is scarce. However, even in large training sets our method obtains a significant improvement. In most cases, though not all, a larger number of augmented observations obtained a larger improvement. When the number of augmented rows was neglible compared to the original training set (e.g., 500 augmented rows compared to 50k training rows) the improvement was very small. Noteably, the entire table is positive - both means and medians, for all three AutoDL models, in all 75 configurations. Figure \ref{fig:guo} presents the results for one single dataset; the figures for all ten datasets are presented in the appendix.

\begin{table*}[]
\caption{DL improvement by augmentation as a function of train size and number of synthesized rows.}
\label{tab:large_datasets}
\small
\centering
\setlength{\tabcolsep}{3pt}
\begin{tabular}{cccccc}
\toprule
\multicolumn{6}{c}{\textbf{H2O mean (median)}} \\
\cmidrule(lr){1-6}
\multicolumn{1}{c}{} & \multicolumn{5}{c}{\textbf{Train size}} \\
\cmidrule(lr){2-6}
\textbf{\# of synthesized rows} & \textbf{500} & \textbf{1,000} & \textbf{5,000} & \textbf{10,000} & \textbf{50,000 or 80\%} \\
\midrule
500      & 5.03\% (3.13\%)   & 3.34\% (2.88\%)  & 0.80\% (0.54\%)   & 0.05\% (-0.14\%)   & 0.12\% (0.08\%)    \\
1,000    & 8.77\% (4.73\%)    & 4.68\% (3.51\%)   & 1.41\% (1.24\%)   & 0.50\% (0.41\%)     & 0.35\% (0.20\%)    \\
5,000    & 8.99\% (6.93\%)   & 8.77\% (6.09\%)  & 3.83\% (2.83\%)  & 3.68\% (3.30\%)    & 1.47\% (1.32\%)    \\
10,000   & 7.76\% (7.03\%)   & 6.80\% (7.32\%)  & 5.53\% (4.40\%)  & 5.20\% (4.35\%)     & 2.49\% (1.95\%)    \\
50,000   & 12.70\% (12.43\%) & 13.87\% (14.54\%) & 13.90\% (11.93\%) & 10.99\% (9.05\%) & 2.68\% (1.23\%)    \\
\midrule
\multicolumn{6}{c}{\textbf{AutoKeras mean (median)}} \\
\cmidrule(lr){1-6}
\multicolumn{1}{c}{} & \multicolumn{5}{c}{\textbf{Train size}} \\
\cmidrule(lr){2-6}
\textbf{\# of synthesized rows} & \textbf{500} & \textbf{1,000} & \textbf{5,000} & \textbf{10,000} & \textbf{50,000 or 80\%} \\
\midrule
500      & 37.03\% (5.31\%)  & 10.69\% (2.35\%)   & 3.11\% (3.88\%)  & 22.48\% (6.90\%)   & 1.12\% (0.10\%)    \\
1,000    & 21.66\% (7.05\%)  & 48.24\% (6.55\%)  & 5.63\% (4.69\%)  & 37.52\% (6.19\%)   & 1.78\% (0.63\%)    \\
5,000    & 60.37\% (18.08\%) & 39.74\% (13.80\%)  & 8.71\% (7.00\%)  & 42.15\% (8.71\%)   & 35.54\% (6.65\%)  \\
10,000   & 85.12\% (20.32\%) & 63.28\% (16.70\%)  & 7.82\% (6.37\%)  & 43.43\% (6.53\%)   & 33.56\% (7.02\%)  \\
50,000   & 63.34\% (23.14\%)  & 40.03\% (16.36\%) & 9.24\% (9.04\%) & 45.93\% (7.39\%)  & 29.95\% (4.72\%)    \\
\midrule
\multicolumn{6}{c}{\textbf{AutoGluon mean (median)}} \\
\cmidrule(lr){1-6}
\multicolumn{1}{c}{} & \multicolumn{5}{c}{\textbf{Train size}} \\
\cmidrule(lr){2-6}
\textbf{\# of synthesized rows} & \textbf{500} & \textbf{1,000} & \textbf{5,000} & \textbf{10,000} & \textbf{50,000 or 80\%} \\
\midrule
500      & 8.48\% (3.53\%)  & 5.57\% (1.92\%)   & 3.15\% (2.01\%)  & 2.23\% (1.31\%)    & 3.21\% (1.47\%)    \\
1,000    & 8.12\% (3.13\%)  & 6.43\% (2.79\%)    & 3.25\% (2.29\%)  & 2.72\% (1.56\%)    & 2.74\% (0.80\%)    \\
5,000    & 11.65\% (6.05\%) & 9.06\% (3.30\%)   & 3.95\% (1.56\%)  & 3.15\% (2.12\%)    & 3.00\% (1.59\%)     \\
10,000   & 9.58\% (5.81\%)  & 9.01\% (3.61\%)    & 4.18\% (1.69\%)  & 3.13\% (1.91\%)     & 2.75\% (1.01\%)    \\
50,000   & 8.12\% (4.40\%)  & 4.62\% (3.25\%)   & 2.96\% (2.05\%)  & 2.29\% (1.93\%)    & 1.39\% (-0.03\%)    \\
\bottomrule
\end{tabular}
\end{table*}

\begin{figure*}[htp]
    \centering
    \hspace*{-0.15\textwidth}
    \includegraphics[width=18cm]{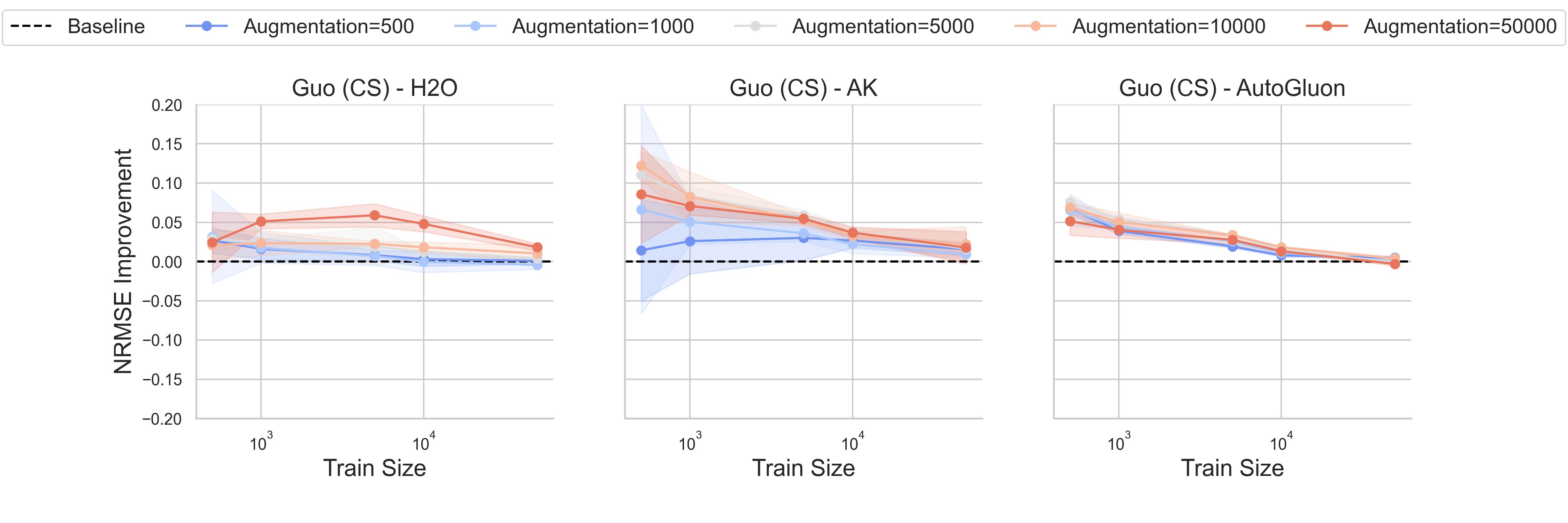}
    \caption{Relative RMSE improvement as a function of training size and number of augmented rows. A summary of ten iterations for the Guo (CS) dataset with varying numbers of augmented rows compared to a baseline without augmentation. The results for all ten datasets are presented in the appendix.}
    \label{fig:guo}
\end{figure*}

\subsection{Augmentation or distillation?}
\label{subsec:aug_or_dist}

\begin{figure*}
    \centering
    \includegraphics[width=0.85\textwidth]{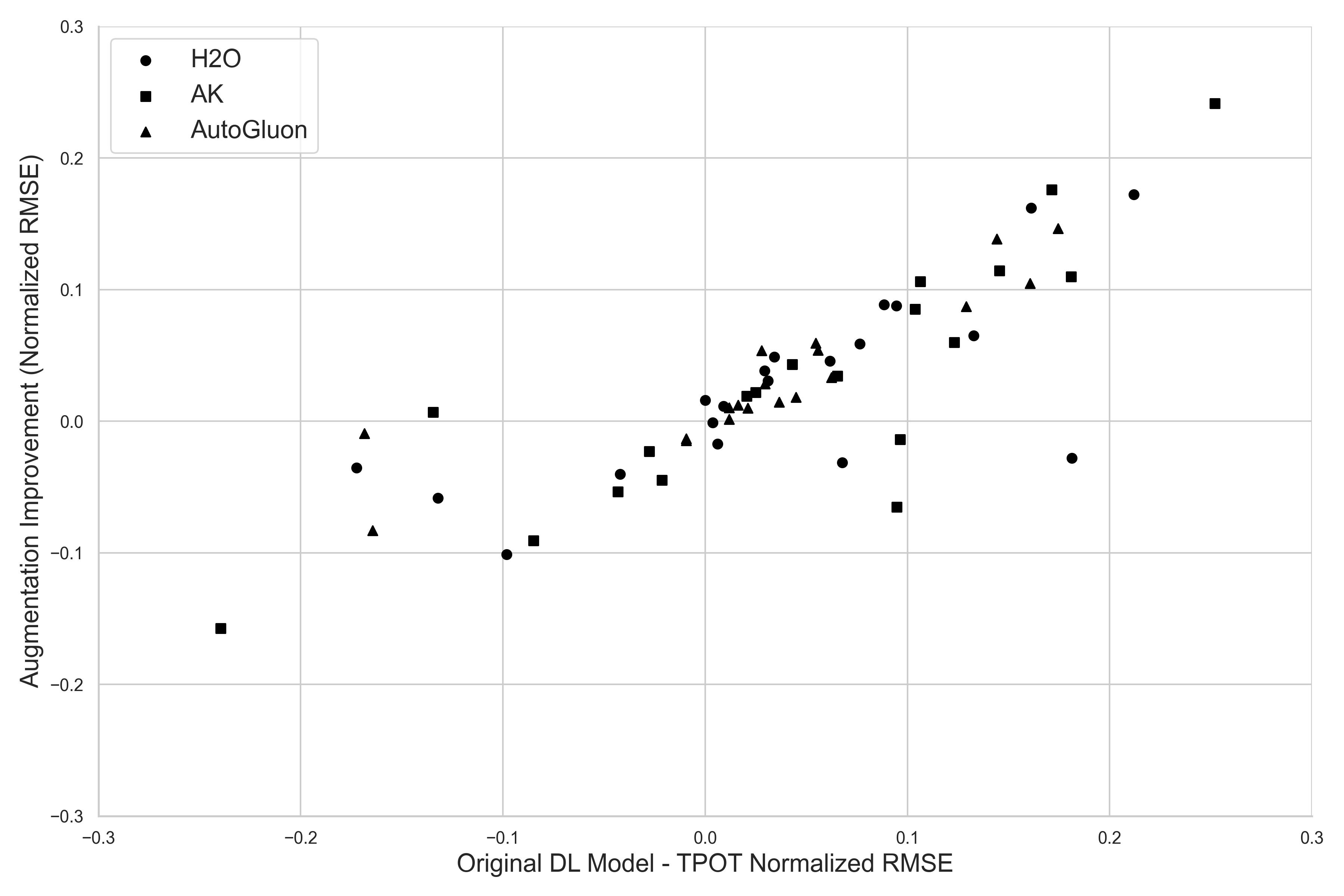}
    \caption{A breakdown of the improvement by augmentation as a function of the advantage of the TPOT model compared to the original DL model. Positive values on the horizontal axis represent an advantage of the TPOT model compared to each DL; positive values on the vertical axis represent an improvement by the augmentation. Each observation represents the mean of one dataset.}
    \label{fig:scatter}
\end{figure*}

We now examine whether the performance improvement originated from the augmentation or was a result of knowledge distillation from the TPOT model. Figure \ref{fig:scatter} shows the performance of the AutoDL models compared to the TPOT model (horizontal axis) and the improvement obtained through augmentation (vertical axis). This improvement is measured by the difference between the AutoDL with augmentation and its baseline without augmentation.

Notably, when the TPOT model outperformed the AutoDL model (positive values on the horizontal axis), the improvement from augmentation was significantly higher. This suggests that knowledge distillation played a pivotal role in the success of our method. However, distillation alone does not fully explain the improvement by our method; as we have shown in Table \ref{tab:large_datasets}, increasing the number of augmented observations is positively correlated with the enhancement in performance. We also explicitly test if knowledge distillation by itself is sufficient, and find that it obtains lesser results than the proposed method. These findings indicate that both augmentation and knowledge distillation played pivotal roles in our method.

\section{Discussion}
\label{sec:discussion}
In this study, we explored the effectiveness of our DA method, which leverages AutoML frameworks to synthesize data, in enhancing the performance of DL models on tabular regression tasks. We compared our method to state-of-the-art DA methods such as C-MixUp \cite{c-mixup} and ADA \cite{base_paper}, and also to naive DA by adding noise, and a baseline without DA. Our results demonstrate a consistent and significant improvement in model performance across diverse datasets with a clear advantage over previous DA methods, thereby highlighting the potential of DA to bridge the gap between DL and classical ML models in tabular data settings.

The results presented in Figure \ref{fig:DL_results} illustrate that the performance of DL models improves as the size of the training dataset increases. This is consistent with the well-known behavior of DL models benefiting from larger datasets due to their ability to learn more complex patterns \cite{dl_size_analysis}. Nonetheless, the extent and consistency of this improvement across multiple datasets and configurations, as shown in our study, provide robust empirical evidence supporting this observation. Moreover, a joint analysis of Table \ref{tab:large_datasets} and Figure \ref{fig:relative_improvement} reveal that DA techniques are particularly beneficial in scenarios where the original data is scarce, as also picked by other studies \cite{da_small_dataset_better}. That said, the improvement in performance was observed not only in small datasets but also in larger ones, albeit to a lesser extent. This suggests that while our DA method is most crucial for small datasets, it still offers value in enhancing the robustness and generalizability of DL models even when more substantial amounts of data are available.

Our comparative analysis across different DA techniques, including C-Mixup and ADA, indicates varied levels of effectiveness. The C-Mixup method, which adjusts sampling probabilities based on label similarities, showed significant promise, particularly in maintaining the integrity of the generated data points and their corresponding labels. This addresses a key challenge in DA for regression tasks where minor changes in the input space can result in significant changes in the output space, unlike classification tasks \cite{dl_regression_harder}. However, the obtained results highlight that our proposed DA method outperforms existing techniques like ADA and C-Mixup, as evidenced in Table \ref{tab:benchmark}. This superior performance can be attributed to our two-step approach, where new source space samples are generated and subsequently, using AutoML, an approximated target feature is computed. An intriguing aspect of our study was the evaluation of whether performance improvements were primarily due to augmentation or knowledge distillation from the TPOT model. As shown in Figure \ref{fig:scatter}, the effectiveness of our augmentation method was strongly dependent on the performance of the AutoML model used for DA. This demonstrates that both the augmentation itself and knowledge distillation contributed to the observed improvement in performance.

\textbf{Limitations and future work.} Despite the promising results, this study has several limitations that warrant consideration. The primary limitation of this method is its dependence on an ML model for data augmentation. While this may not pose significant issues for tabular data predictions, as discussed in this paper, it may present greater challenges in other fields such as computer vision. Furthermore, even within the realm of tabular data, ML models can exhibit low performance. We have demonstrated that when the ML model used for data augmentation performs poorly, it can negatively impact the performance of the DL model. Additionally, this method requires the training of two separate models, thereby increasing computational effort (it is worth noting that additional augmentation methods like ADA \cite{base_paper} also require extensive computation time). Finally, although we included three prominent AutoML frameworks (H2O, AutoKeras, and AutoGluon) and tested them over 30 datasets, the results may vary with other frameworks and datasets not covered in this study. Future work should address these limitations by exploring a wider range of datasets, tasks, and DL frameworks.

\textbf{Conclusions.} Taken jointly, our study demonstrates that DA can significantly enhance the performance of DL models on tabular data regression tasks. By addressing some common limitations in tabular data, such as scarcity, DA techniques provide a powerful tool to improve model performance and generalizability. The empirical evidence from our extensive experiments suggests that these techniques can help bridge the performance gap between DL and classical ML models in this domain. Future research should continue to explore and refine these methods.

\bibliography{biblio}
\bibliographystyle{IEEEtran}

\section{Supplementary Material}

Table \ref{tab:datasets} summarizes the datasets used in the research.

\begin{table*}[!hb]
\centering
\caption{Descriptive statistics of the datasets used in the study.}
\label{tab:datasets}
\begin{tabular}{@{}llllllll@{}}
\toprule
\textbf{\#} & \textbf{Dataset}                                                       & \textbf{Reference} & \textbf{Section}             & \textbf{\#rows} & \textbf{\#cols} & \textbf{Target mean} & \textbf{Target STD} \\ \midrule
A          & Medical costs                                                          & \cite{medical}     & \ref{subsec:twenty_datasets} & 1338            & 7              & 13270         & 12110        \\
B          & Wine                                                                   & \cite{wine}        & \ref{subsec:twenty_datasets} & 1143            & 12             & 5.66          & 0.81         \\
C          & Advertising                                                            & \cite{advertising} & \ref{subsec:twenty_datasets} & 200             & 4              & 15.13         & 5.28         \\
D          & Real estate                                                            & \cite{real_estate} & \ref{subsec:twenty_datasets} & 414             & 7              & 37.98         & 13.61        \\
E          &  Concrete strength          & \cite{concrete}    & \ref{subsec:twenty_datasets} & 1030            & 9              & 35.82         & 16.71        \\
F          & Cellphone                                                              & \cite{cellphone}   & \ref{subsec:twenty_datasets} & 161             & 13             & 2215.6        & 768.19       \\
G          & Airfoil                                                            & \cite{airfoil} & \ref{subsec:twenty_datasets} & 1503           & 6             & 124.84        & 125.72       \\
H          & Bike daily                                                             & \cite{bike_rental} & \ref{subsec:twenty_datasets} & 731             & 12             & 4504          & 1937         \\
I          & Admission chance                                                       & \cite{admission}   & \ref{subsec:twenty_datasets} & 500             & 8              & 0.72          & 0.14         \\
J         & Huang (FS)                                                             & \cite{huang}       & \ref{subsec:twenty_datasets} & 114             & 10             & 10.26         & 2.64         \\
K         & Huang (CS)                                                             & \cite{huang}       & \ref{subsec:twenty_datasets} & 114             & 10             & 73.13         & 29.74        \\
L         & Su (1)                                                                 & \cite{su}          & \ref{subsec:twenty_datasets} & 122             & 8              & 8.56          & 3.38         \\
M         & Su (2)                                                                 & \cite{su}          & \ref{subsec:twenty_datasets} & 122             & 8              & 12.05         & 4.05         \\
N         & Bachir                                                                 & \cite{bachir}      & \ref{subsec:twenty_datasets} & 112             & 4              & 40.27         & 19.59        \\
O         & Yin (Fmax)                                                             & \cite{yin}         & \ref{subsec:twenty_datasets} & 925             & 12             & 0.30          & 0.44         \\
P         & Yin (ifss)                                                             & \cite{yin}         & \ref{subsec:twenty_datasets} & 925             & 12             & 34.00         & 28.58        \\
Q         & Matbench                                                               & \cite{matbench}    & \ref{subsec:twenty_datasets} & 312             & 15             & 1421          & 302          \\
R         & Atici                                                                  & \cite{Atici}       & \ref{subsec:twenty_datasets} & 140             & 4              & 20.43         & 6.53         \\
S         & Xiong (height)             & \cite{xiong}       & \ref{subsec:twenty_datasets} & 43              & 5              & 2.87          & 0.36         \\
T         & Xiong (width)                                                          & \cite{xiong}       & \ref{subsec:twenty_datasets} & 43              & 5              & 9.00          & 1.23         \\ \midrule
A1         & Guo (CS)                                                               & \cite{guo}         & \ref{subsec:large_datasets}  & 63162           & 28             & 365.83        & 123.11       \\
A2         & Guo (TS)                                                               & \cite{guo}         & \ref{subsec:large_datasets}  & 63162           & 28             & 455.77        & 123.43       \\
A3         & Guo (EL)                                                               & \cite{guo}         & \ref{subsec:large_datasets}  & 63162           & 28             & 37.33         & 9.11         \\
A4         & Abalone                                                                & \cite{abalone}     & \ref{subsec:large_datasets}  & 94792           & 9              & 9.71          & 3.18         \\
A5         & Air BNB                                                                & \cite{airbnb}      & \ref{subsec:large_datasets}  & 14998           & 17             & 130.05        & 63.82        \\
A6         & Electricity Prices                                                     & \cite{electricity_prices} & \ref{subsec:large_datasets}  & 38014           & 17             & 64.14         & 55.45        \\
A7         & Energy                                                                 & \cite{energy}      & \ref{subsec:large_datasets}  & 19735           & 28             & 97.69         & 102.52       \\
A8         & Gems                                                                   & \cite{gems}        & \ref{subsec:large_datasets}  & 193573          & 10             & 3969.16       & 4034.37      \\
A9         & Maintenance                                                            & \cite{maintenance} & \ref{subsec:large_datasets}  & 39414           & 5              & 111.49        & 74.14        \\
A10         & Wind Power                                                            & \cite{wind}        & \ref{subsec:large_datasets}  & 43800          & 9              & 0.41         & 0.29         \\ \bottomrule
\end{tabular}
\end{table*}

Tables \ref{tab:h2o_errors}, \ref{tab:AK_errors}, and \ref{tab:gluon_errors} present the errors bars of the results originally presented in Table \ref{tab:large_datasets} for the H2O, AutoKeras, and AutoGluon models, respectively. 
Table \ref{table:first_experiment_p_values} shows the results of a multivariate analysis of variance (MANOVA) test between the best-performing result and the other results, following Table \ref{tab:benchmark}, to explore if the improvement of the best-performing method for each dataset and data augmentation method is statistically significant.

\begin{table*}[!ht]
\caption{A comparison between ADA, C-mixup, naive noise, and our method in 20 datasets with the H2O library. The best augmentation method in each dataset (row) is in bold. Error bars represent 95\% confidence level.}
\label{tab:h2o_errors}
\small
\centering
\setlength{\tabcolsep}{3pt}
\begin{tabular}{@{}lccccc@{}}
\toprule
\textbf{Dataset} & \textbf{Ours}           & \textbf{ADA}       & \textbf{C-mixup} & \textbf{Noise}        & \textbf{Baseline}  \\
\midrule
A                & \textbf{4495.29±246.21} & 6055.38±216.77     & 5824.46±222.95   & 5963.02±226.27        & 5544.78±122.35     \\
B                & \textbf{0.61±0.02}      & 0.69±0.01          & 0.70±0.02        & 0.67±0.02             & 0.74±0.01          \\
C                & \textbf{1.13±0.10}      & 1.63±0.10          & 1.94±0.10        & 1.85±0.08             & 1.51±0.07          \\
D                & \textbf{6.34±0.43}      & 9.58±1.02          & 11.48±0.97       & 9.02±0.80             & 8.62±0.62          \\
E                & \textbf{4.70±0.21}      & 5.20±0.28          & 5.09±0.35        & 5.43±0.19             & 5.12±0.14          \\
F                & 156.95±22.94            & 133.20±25.12       & 131.92±54.76     & \textbf{131.45±33.37} & 150.06±9.97        \\
G                & 1.68±0.06               & \textbf{1.53±0.05} & 1.73±0.09        & 1.56±0.04             & 1.69±0.04          \\
H                & \textbf{636.34±22.36}   & 754.75±32.38       & 818.49±33.41     & 711.85±34.59          & 721.42±18.95       \\
I                & \textbf{0.06±0.00}      & 0.08±0.00          & 0.08±0.00        & 0.07±0.00             & 0.08±0.00          \\
J                & 1.45±0.14               & 1.48±0.25          & 1.66±0.20        & \textbf{1.29±0.21}    & 1.66±0.09          \\
K                & 10.78±3.09              & 10.77±1.40         & 12.01±0.88       & \textbf{9.57±1.18}    & 10.66±0.95         \\
L                & 1.72±0.65               & \textbf{1.30±0.15} & 1.91±0.20        & 1.48±0.14             & 1.69±0.18          \\
M                & \textbf{1.54±0.02}      & 1.55±0.20          & 1.59±0.06        & 1.80±0.25             & 1.62±0.07          \\
N                & 4.61±0.91               & \textbf{2.97±0.98} & 2.57±0.20        & 4.89±1.63             & 4.37±0.63          \\
O                & 0.04±0.03               & \textbf{0.02±0.00} & 0.02±0.00        & 0.02±0.00             & 0.02±0.00          \\
P                & 6.73±2.44               & \textbf{4.38±0.30} & 4.09±0.17        & 5.30±0.28             & 5.96±0.21          \\
Q                & \textbf{139.93±16.97}   & 186.06±29.75       & 152.95±13.05     & 164.97±32.17          & 159.00±11.68       \\
R                & \textbf{2.24±0.36}      & 3.08±0.31          & 3.45±0.44        & 2.96±0.29             & 3.17±0.19          \\
S                & 0.13±0.03               & 0.09±0.02          & 0.16±0.01        & 0.18±0.03             & \textbf{0.09±0.01} \\
T                & 0.30±0.08               & 0.32±0.05          & 0.51±0.14        & 0.48±0.09             & \textbf{0.26±0.02} \\
\midrule
\textbf{\# best}  & \textbf{10}             & \textbf{2}         & \textbf{3}       & \textbf{3}            & \textbf{2}        \\
\bottomrule
\end{tabular}
\end{table*}

\begin{table*}[!ht]
\caption{A comparison between ADA, C-mixup, naive noise, and our method in 20 datasets with the AutoKeras library. The best augmentation method in each dataset (row) is in bold. Error bars represent 95\% confidence level.}
\label{tab:AK_errors}
\small
\centering
\setlength{\tabcolsep}{3pt}
\begin{tabular}{@{}lccccc@{}}
\toprule
\textbf{Dataset} & \textbf{Ours}           & \textbf{ADA}       & \textbf{C-mixup}   & \textbf{Noise}        & \textbf{Baseline}  \\
\midrule
A                & \textbf{4443.80±256.64} & 5686.24±361.09     & 9513.30±1693.93    & 6043.68±180.65        & 6516.56±96.90      \\
B                & \textbf{0.64±0.02}      & 0.84±0.04          & 0.73±0.02          & 0.84±0.04             & 0.73±0.01          \\
C                & \textbf{1.20±0.11}      & 1.78±0.30          & 1.99±0.15          & 1.78±0.12             & 1.59±0.08          \\
D                & \textbf{6.52±0.44}      & 10.86±1.01         & 12.38±1.02         & 9.82±1.07             & 8.20±0.52          \\
E                & \textbf{4.96±0.19}      & 5.34±0.22          & 5.01±0.35          & 5.82±0.27             & 6.85±0.17          \\
F                & 156.77±23.54            & 134.47±35.60       & 149.68±30.69       & \textbf{131.26±34.05} & 139.36±16.32       \\
G                & 2.05±0.05               & 2.02±0.03          & 2.22±0.09          & \textbf{1.78±0.04}    & 5.44±0.09          \\
H                & \textbf{690.10±32.34}   & 904.43±55.76       & 966.64±147.54      & 828.86±27.45          & 869.91±23.10       \\
I                & \textbf{0.06±0.00}      & 0.10±0.00          & 0.08±0.00          & 0.08±0.00             & 0.08±0.00          \\
J                & 1.48±0.14               & 1.56±0.24          & 1.77±0.19          & 1.52±0.17             & \textbf{1.37±0.06} \\
K                & 11.30±3.28              & 11.15±1.75         & 11.77±0.90         & \textbf{9.97±1.01}    & 10.27±1.02         \\
L                & 1.69±0.60               & 1.49±0.33          & 2.30±0.37          & 1.71±0.27             & \textbf{1.37±0.12} \\
M                & 1.57±0.05               & \textbf{1.54±0.16} & 1.58±0.04          & 1.74±0.23             & 2.20±0.19          \\
N                & 4.56±1.07               & 3.07±1.15          & \textbf{2.95±0.36} & 4.98±1.76             & 3.83±0.60          \\
O                & 0.04±0.02               & 0.02±0.00          & \textbf{0.02±0.00} & 0.03±0.00             & 0.04±0.01          \\
P                & 8.46±3.35               & \textbf{5.25±0.55} & 5.43±0.34          & 5.26±0.32             & 6.85±0.61          \\
Q                & \textbf{152.79±13.88}   & 214.75±53.02       & 189.53±21.20       & 180.89±30.12          & 167.69±16.69       \\
R                & \textbf{2.60±0.35}      & 3.68±0.31          & 3.82±0.47          & 3.31±0.31             & 3.47±0.20          \\
S                & 0.13±0.03               & \textbf{0.08±0.02} & 0.14±0.02          & 0.21±0.04             & 0.10±0.01          \\
T                & 0.31±0.09               & \textbf{0.30±0.06} & 0.40±0.08          & 0.54±0.10             & 0.33±0.03          \\
\midrule
\textbf{\# best}  & \textbf{9}             & \textbf{4}         & \textbf{2}         & \textbf{3}            & \textbf{2}        \\
\bottomrule
\end{tabular}
\end{table*}

\begin{table*}[!ht]
\caption{A comparison between ADA, C-mixup, naive noise, and our method in 20 datasets with the AutoGluon library. The best augmentation method in each dataset (row) is in bold. Error bars represent 95\% confidence level.}
\label{tab:gluon_errors}
\small
\centering
\setlength{\tabcolsep}{3pt}
\begin{tabular}{@{}lccccc@{}}
\toprule
\textbf{Dataset} & \textbf{Ours}         & \textbf{ADA}       & \textbf{C-mixup}      & \textbf{Noise}          & \textbf{Baseline}  \\
\midrule
A                & 4546.54±272.38        & 4658.89±398.78     & 4653.65±232.55        & \textbf{4326.46±280.65} & 4640.12±154.19     \\
B                & \textbf{0.61±0.02}    & 0.65±0.02          & 0.67±0.01             & 0.69±0.02               & 0.64±0.01          \\
C                & \textbf{1.18±0.11}    & 1.53±0.13          & 1.93±0.18             & 1.82±0.09               & 1.63±0.07          \\
D                & \textbf{6.25±0.46}    & 8.33±0.86          & 10.84±1.24            & 8.82±0.93               & 7.50±0.52          \\
E                & \textbf{4.81±0.22}    & 5.32±0.24          & 4.94±0.16             & 5.43±0.20               & 5.25±0.15          \\
F                & 159.25±24.89          & 142.44±27.90       & \textbf{130.30±19.30} & 142.65±40.97            & 178.05±15.34       \\
G                & 2.00±0.07             & 2.01±0.02          & 2.09±0.08             & \textbf{1.67±0.04}      & 2.00±0.05          \\
H                & \textbf{661.45±27.38} & 790.54±54.60       & 774.75±30.76          & 685.94±55.34            & 665.63±18.05       \\
I                & \textbf{0.06±0.00}    & 0.07±0.00          & 0.07±0.00             & 0.08±0.00               & 0.06±0.00          \\
J                & 1.46±0.18             & 1.46±0.15          & 1.53±0.12             & \textbf{1.38±0.27}      & 1.61±0.09          \\
K                & 10.49±3.26            & 10.98±1.56         & 12.36±1.29            & \textbf{9.30±1.23}      & 11.70±1.02         \\
L                & 1.69±0.46             & 1.62±0.25          & \textbf{1.53±0.10}    & 2.15±0.55               & 1.75±0.22          \\
M                & 1.54±0.03             & 1.68±0.19          & 1.53±0.03             & 1.71±0.21               & \textbf{1.51±0.09} \\
N                & 4.67±0.89             & 3.47±1.13          & \textbf{2.98±0.29}    & 5.74±2.34               & 5.23±0.62          \\
O                & 0.03±0.01             & 0.03±0.01          & 0.04±0.00             & \textbf{0.02±0.00}      & 0.02±0.00          \\
P                & 4.20±0.22             & \textbf{4.10±0.24} & 4.55±0.11             & 4.78±0.27               & 4.14±0.13          \\
Q                & 128.03±15.94          & 130.53±14.05       & \textbf{125.50±13.59} & 158.32±26.11            & 165.29±12.66       \\
R                & \textbf{2.46±0.33}    & 3.18±0.36          & 3.55±0.50             & 3.15±0.31               & 3.35±0.17          \\
S                & 0.13±0.02             & \textbf{0.10±0.02} & 0.14±0.01             & 0.13±0.02               & 0.21±0.04          \\
T                & 0.37±0.11             & 0.37±0.07          & \textbf{0.34±0.03}    & 0.48±0.09               & 0.43±0.08          \\
\midrule
\textbf{\# best}  & \textbf{7}           & \textbf{2}         & \textbf{5}            & \textbf{5}              & \textbf{1}        \\
\bottomrule
\end{tabular}
\end{table*}

\begin{table*}[!ht]
\centering
\caption{The p-value of a multivariate analysis of variance (MANOVA) test between the best-performing result and the other results.}
\begin{tabular}{lcccc}
\hline 	\textbf{DS} & 	\textbf{H2O} & 	\textbf{AutoKeras} & 	\textbf{AutoGluon} \\ \hline 
 A & <0.05 & <0.05 & <0.01 \\  
 B & <0.05 & <0.05 & <0.05 \\  
 C & <0.05 & <0.05 & <0.05 \\  
 D & <0.01 & <0.01 & <0.05 \\  
 E & <0.05 & <0.05 & <0.05 \\  
 F & >0.05 & >0.05 & >0.05 \\  
 G & <0.05 & <0.05 & <0.05 \\  
 H & <0.01 & <0.01 & <0.01 \\  
 I & >0.05 & >0.05 & >0.05 \\  
 J & <0.05 & <0.01 & <0.01 \\  
 K & <0.01 & <0.05 & <0.05 \\  
 L & <0.01 & <0.05 & <0.05 \\  
 M & >0.05 & >0.05 & <0.05 \\  
 N & <0.01 & <0.01 & <0.01 \\  
 O & >0.05 & >0.05 & >0.05 \\  
 P & >0.05 & >0.05 & >0.05 \\  
 Q & <0.05 & >0.05 & <0.05 \\  
 R & <0.01 & <0.05 & <0.05 \\  
 S & <0.05 & <0.05 & <0.01 \\  
 T & >0.05 & >0.05 & >0.05 \\  \hline 
\end{tabular}
\label{table:first_experiment_p_values}
\end{table*}

\begin{figure*}[!htp]
    \centering
    \includegraphics[width=11.5cm]{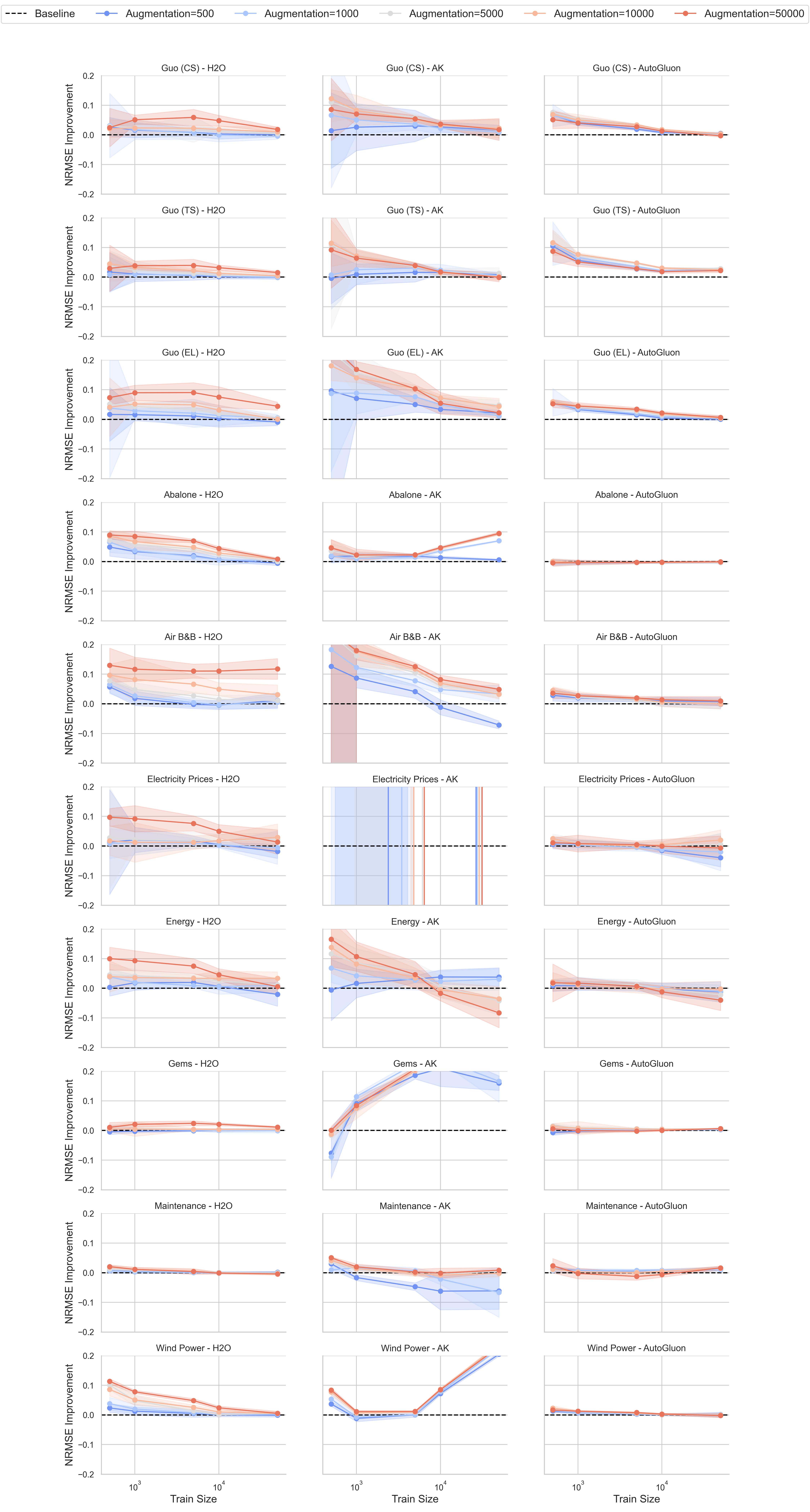}
    \caption{Relative RMSE improvement as a function of training size and number of augmented rows. A summary of ten iterations for ten datasets with varying numbers of augmented rows compared to a baseline without augmentation.}
    \label{fig:relative_improvement}
\end{figure*}

Figure \ref{fig:relative_improvement} presents the relative improvement gained by augmentation as a function of the original training size and the number of augmented rows. The figure is parallel to Figure \ref{fig:guo}, but for all ten large datasets. Its results were also summarized in Table \ref{tab:large_datasets}.

Table \ref{tab:large_errors} is equivalent to Table \ref{tab:large_datasets} with error bars at the 95\% confidence level.

\begin{table*}[]
\caption{DL improvement by augmentation as a function of train size and number of synthesized rows.}
\label{tab:large_errors}
\small
\centering
\setlength{\tabcolsep}{3pt}
\begin{tabular}{cccccc}
\toprule
\multicolumn{6}{c}{\textbf{H2O mean (median)}} \\
\cmidrule(lr){1-6}
\multicolumn{1}{c}{} & \multicolumn{5}{c}{\textbf{Train size}} \\
\cmidrule(lr){2-6}
\textbf{\# of synthesized rows} & \textbf{500} & \textbf{1,000} & \textbf{5,000} & \textbf{10,000} & \textbf{50,000 or 80\%} \\
\midrule
500      & 5.03\% ± 0.01\%   & 3.34\% ± 0.00\%  & 0.80\% ± 0.00\%   & 0.05\% ± 0.00\%   & 0.12\% ± 0.00\%    \\
1,000    & 8.77\% ± 0.02\%    & 4.68\% ± 0.01\%   & 1.41\% ± 0.00\%   & 0.50\% ± 0.00\%     & 0.35\% ± 0.00\%    \\
5,000    & 8.99\% ± 0.01\%   & 8.77\% ± 0.01\%  & 3.83\% ± 0.00\%  & 3.68\% ± 0.00\%    & 1.47\% ± 0.00\%    \\
10,000   & 7.76\% ± 0.01\%   & 6.80\% ± 0.01\%  & 5.53\% ± 0.00\%  & 5.20\% ± 0.00\%     & 2.49\% ± 0.00\%    \\
50,000   & 12.70\% ± 0.01\% & 13.87\% ± 0.00\% & 13.90\% ± 0.01\% & 10.99\% ± 0.01\% & 2.68\% ± 0.00\%    \\
\midrule
\multicolumn{6}{c}{\textbf{AutoKeras mean (median)}} \\
\cmidrule(lr){1-6}
\multicolumn{1}{c}{} & \multicolumn{5}{c}{\textbf{Train size}} \\
\cmidrule(lr){2-6}
\textbf{\# of synthesized rows} & \textbf{500} & \textbf{1,000} & \textbf{5,000} & \textbf{10,000} & \textbf{50,000 or 80\%} \\
\midrule
500      & 37.03\% ± 0.08\%  & 10.69\% ± 0.04\%   & 3.11\% ± 0.01\%  & 22.48\% ± 0.05\%   & 1.12\% ± 0.01\%    \\
1,000    & 21.66\% ± 0.04\%  & 48.24\% ± 0.29\%  & 5.63\% ± 0.00\%  & 37.52\% ± 0.08\%   & 1.78\% ± 0.02\%    \\
5,000    & 60.37\% ± 0.09\% & 39.74\% ± 0.11\%  & 8.71\% ± 0.00\%  & 42.15\% ± 0.08\%   & 35.54\% ± 0.06\%  \\
10,000   & 85.12\% ± 0.12\% & 63.28\% ± 0.21\%  & 7.82\% ± 0.01\%  & 43.43\% ± 0.08\%   & 33.56\% ± 0.06\%  \\
50,000   & 63.34\% ± 0.10\%  & 40.03\% ± 0.08\% & 9.24\% ± 0.00\% & 45.93\% ± 0.07\%  & 29.95\% ± 0.05\%    \\
\midrule
\multicolumn{6}{c}{\textbf{AutoGluon mean (median)}} \\
\cmidrule(lr){1-6}
\multicolumn{1}{c}{} & \multicolumn{5}{c}{\textbf{Train size}} \\
\cmidrule(lr){2-6}
\textbf{\# of synthesized rows} & \textbf{500} & \textbf{1,000} & \textbf{5,000} & \textbf{10,000} & \textbf{50,000 or 80\%} \\
\midrule
500      & 8.48\% ± 0.01\%  & 5.57\% ± 0.01\%   & 3.15\% ± 0.00\%  & 2.23\% ± 0.00\%    & 3.21\% ± 0.00\%    \\
1,000    & 8.12\% ± 0.01\%  & 6.43\% ± 0.01\%    & 3.25\% ± 0.00\%  & 2.72\% ± 0.00\%    & 2.74\% ± 0.00\%    \\
5,000    & 11.65\% ± 0.01\% & 9.06\% ± 0.01\%   & 3.95\% ± 0.00\%  & 3.15\% ± 0.00\%    & 3.00\% ± 0.00\%     \\
10,000   & 9.58\% ± 0.01\%  & 9.01\% ± 0.01\%    & 4.18\% ± 0.00\%  & 3.13\% ± 0.00\%     & 2.75\% ± 0.00\%    \\
50,000   & 8.12\% ± 0.01\%  & 4.62\% ± 0.01\%   & 2.96\% ± 0.00\%  & 2.29\% ± 0.00\%    & 1.39\% ± 0.00\%    \\
\bottomrule
\end{tabular}
\end{table*}

\end{document}